\documentclass[11pt,oneside,english]{amsart}

\usepackage[T1]{fontenc}
\usepackage[latin9]{inputenc}
\usepackage[letterpaper]{geometry}
\usepackage{float}
\usepackage{amsbsy}
\usepackage{amsthm}
\usepackage{amssymb}
\usepackage{graphicx}

\makeatletter

\providecommand{\tabularnewline}{\\}
\floatstyle{ruled}
\newfloat{algorithm}{tbp}{loa}
\providecommand{\algorithmname}{Algorithm}
\floatname{algorithm}{\protect\algorithmname}

\numberwithin{equation}{section}

\usepackage{ifpdf} 
\ifpdf 
 \IfFileExists{lmodern.sty}{\usepackage{lmodern}}{}
\fi 
\usepackage{color}

\usepackage{fancyhdr}
\fancypagestyle{plain}{\fancyfoot{}\fancyhead[C]{\text{\thepage}}}
\usepackage{graphicx}

\usepackage{hyperref}
\makeatother

\usepackage{babel}
\begin{document}

\title{Discrete Linear-Complexity Reinforcement Learning in Continuous Action
Spaces For Q-Learning Algorithms }

\author{Peyman Tavallali$^{1,*}$, Gary B. Doran Jr.$^{1}$, Lukas Mandrake$^{1}$}
\begin{abstract}
In this article, we sketch an algorithm that extends the Q-learning
algorithms to the continuous action space domain. Our method is based
on the discretization of the action space. Despite the commonly used
discretization methods, our method does not increase the discretized
problem dimensionality exponentially. We will show that our proposed
method is linear in complexity when the discretization is employed.
The variant of the Q-learning algorithm presented in this work, labeled
as Finite Step Q-Learning (FSQ), can be deployed to both shallow and
deep neural network architectures.
\end{abstract}

\maketitle
$^{1}$Jet Propulsion Laboratory, California Institute of Technology,
Pasadena, CA 91109

$^{*}$Corresponding Author, email: peyman.tavallali@jpl.nasa.gov,
tavallali@gmail.com

\section{Introduction}

\begin{figure}[H]
\includegraphics[scale=0.4]{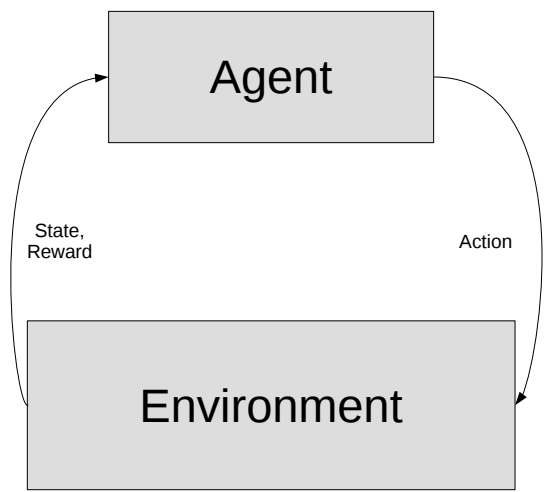}

\caption{RL: In this figure, the feedback closed-loop interaction between an
RL agent and the environment is depicted.}

\label{Fig: RL}
\end{figure}

Reinforcement Learning (RL) is one of the most general forms of Machine
Learning (ML) and Artificial Intelligence (AI) algorithms. RL concerns
with a learning autonomous ``agent'' that interacts with its environment
\cite{sutton1998reinforcement}. In this setting, an agent, in a certain
state, takes either a deterministic or probabilistic action and consequently
receives a deterministic or stochastic reward from the environment
(See Figure \ref{Fig: RL}). RL is of critical importance in AI since
an RL agent is autonomously interacting with the environment and learns
from this interaction by maximizing the reward it achieves. There
is no explicit supervision in RL, but only learning through reward
and punishment.

There has been a significant attention to RL, in recent years, as
RL has shown exemplary capability in solving problems across various
domains; specially when combined with deep neural networks \cite{goodfellow2016deep,henderson2017deep,levine2016end}.
In what comes next, we introduce the notation commonly used in RL.

\subsection{Notation}

The state $s_{t}\in\mathcal{S}$ is what an agent is experiencing
at time $t$. An action, $a_{t}\in\mathcal{A}\left(s_{t}\right)$,
belongs to the set of all possible actions $\mathcal{A}\left(s_{t}\right)$
that the agent can take at time $t$. Generally, we have $\mathcal{S}\subset\mathbb{R}^{n},n\in\mathbb{N}$
and $\mathcal{A}\subset\mathbb{R}^{m},m\in\mathbb{N}$. However, for
a discrete setting we could have $\mathcal{S}\subset\mathbb{N},\,card\left(\mathcal{S}\right)<\infty$
or $\mathcal{A}\subset\mathbb{N},\,card\left(\mathcal{A}\right)<\infty$.
The reward that an agent receives from the environment at any time
instance is $r_{t}$ for $r:\mathcal{S}\times\mathcal{A}\rightarrow\mathbb{R}$.
This is a deterministic notation for the reward $r_{t}$ signal. However,
the reward signal can be a real valued random variable drawn from
a specific distribution. 

It is usually assumed that the agent and the environment follow a
Markov Decision Process (MDP) property. To be more specifice, we define
the MDP as 
\[
\mathcal{M}=\left\{ \mathcal{S},\mathcal{A},\mathcal{P},r\right\} ,
\]
in which $\mathcal{P}$ is the MDP state transition probability kernel
\cite{szepesvari2010algorithms} that assigns to each state-action
pair $\left(s,a\right)\in\mathcal{S}\times\mathcal{A}$ a probability
measure over $\mathcal{S}\times\mathbb{R}$. For a discrete setting
it is defined as a tensor that can be expressed in an element-wise
format as 
\begin{equation}
\mathcal{P}_{ss^{'}}^{a}=\Pr\left\{ s_{t+1}=s^{'}\left|s_{t}=s,a_{t}=a\right.\right\} .\label{eq: Transition_op}
\end{equation}
The Markov property of $\mathcal{M}$, in a discrete setting, imposes
that 
\begin{equation}
\begin{array}{c}
\Pr\left\{ s_{t+1}=s^{'}\right\} =\sum_{s,a}\Pr\left\{ s_{t+1}=s^{'}\left|s_{t}=s,a_{t}=a\right.\right\} \Pr\left\{ a_{t}=a\left|s_{t}=s\right.\right\} \Pr\left\{ s_{t}=s\right\} \end{array}.\label{eq: Markov_prop}
\end{equation}
In other words, $s_{t+1}$ only depends on $s_{t}$ and $a_{t}$. 

The action that an agent takes based on its current state, in the
environment, is called policy or behavior. In (\ref{eq: Markov_prop}),
the policy is defined $\pi\left(s,a\right)=\Pr\left\{ a_{t}=a\left|s_{t}=s\right.\right\} $.
This notation is a probabilistic format of the policy. A deterministic
policy can be expressed as $a=\pi\left(s\right)$. 

Maximizing the current reward $r_{t}$ is not the best policy that
an autonomous agent can take. It is a discounted return (weighted
average) of the current and future rewards that matters. In other
words, the policy $\pi\left(s,a\right)$ should maximize either the
discounted return
\begin{equation}
R_{t}=\sum_{k=0}^{T}\gamma^{k}r_{t+1+k},\label{eq: Discounted Return}
\end{equation}
or a representation of this value. In Equation (\ref{eq: Discounted Return}),
the discount factor $\gamma$ belongs to $\left(0,1\right]$, also
$T$ specifies the end of an episode which could be infinite. 

From a mathematical perspective, since $r_{t}$ could be stochastic,
it is the expectation value of $R_{t}$ that should be maximized.
Hence, an RL agent tries to find a policy $\pi\left(s,a\right)$ that
maximizes all discounted future rewards. To be more specific, an RL
agent must maximize the value function 
\begin{equation}
V^{\pi}\left(s\right)=\mathbb{E}_{\pi}\left\{ R_{t}\left|s_{t}=s\right.\right\} ,\label{eq: Value_fun}
\end{equation}
or the action-value function
\begin{equation}
Q^{\pi}\left(s,a\right)=\mathbb{E}_{\pi}\left\{ R_{t}\left|s_{t}=s,a_{t}=a\right.\right\} ,\label{eq: action_value_fun}
\end{equation}
for a policy $\pi\left(s,a\right)$. These two functions satisfy certain
recursive equations, commonly called Bellman Equations \cite{bellman2013dynamic},
as follow
\begin{equation}
V^{\pi}\left(s\right)=\mathbb{E}_{\pi}\left\{ r_{t+1}+\gamma V^{\pi}\left(s_{t+1}=s^{'}\right)\left|s_{t}=s\right.\right\} ,\label{eq: V_Bellman}
\end{equation}
\begin{equation}
Q^{\pi}\left(s,a\right)=\mathbb{E}_{\pi}\left\{ r_{t+1}+\gamma V^{\pi}\left(s_{t+1}=s^{'}\right)\left|s_{t}=s,a_{t}=a\right.\right\} .\label{eq: Q_Bellman}
\end{equation}
Equations (\ref{eq: V_Bellman}) and (\ref{eq: Q_Bellman}) are recursive
functions that play important roles in designing RL algorithms, and
in specific, the family of Q-learning algorithms \cite{bonarini2009reinforcement,fukao1998q,gaskett1999q,gu2016continuous,hasselt2010double,kobayashi2014q,millan2002continuous,rummery1994line,takahashi1999continuous,van2016deep,watkins1992q}.
These equations are more informative in their optimal value settings.
From an optimization perspective, for an action-value function, an
RL agent tries to find solutions to 
\begin{equation}
Q^{*}\left(s,a\right)=\underset{\pi}{\max}\,\mathbb{E}_{\pi}\left\{ R_{t}\left|s_{t}=s,a_{t}=a\right.\right\} .\label{eq: Optimal_Q}
\end{equation}
This solution coincides with the solution to 
\begin{equation}
V^{*}\left(s\right)=\underset{\pi}{\max}\,\mathbb{E}_{\pi}\left\{ R_{t}\left|s_{t}=s\right.\right\} .\label{eq: Optimal_v}
\end{equation}
Using the Markov property, the optimal Bellman equations \cite{bellman2013dynamic}
can be derived for both value and action-value functions as follows
\begin{equation}
Q^{*}\left(s,a\right)=\mathbb{E}\left\{ r_{t+1}+\gamma V^{*}\left(s_{t+1}=s^{'}\right)\left|s_{t}=s,a_{t}=a\right.\right\} ,\label{eq: Q_Bellman_Optimal}
\end{equation}
\begin{equation}
V^{*}\left(s\right)=\underset{a\in\mathcal{A}\left(s\right)}{\max}Q^{*}\left(s,a\right).\label{eq: V_optimal4Bellman}
\end{equation}
One subtle fact is that, in Equation (\ref{eq: Q_Bellman_Optimal}),
the expectation is not taken over all possible policies. This fact
helps the design of algorithms to have more freedom from a numerical
perspective when the action-value function $Q^{*}\left(s,a\right)$
is being estimated. Soon, we show how these equations are being used
in Q-learning algorithms \cite{bonarini2009reinforcement,fukao1998q,gaskett1999q,gu2016continuous,hasselt2010double,kobayashi2014q,millan2002continuous,rummery1994line,takahashi1999continuous,van2016deep,watkins1992q}.
However, in order to reach that point we need to explain a quick histroy
of the RL algorithms in general.

\subsection{RL Algorithms}

\begin{figure}[H]
\includegraphics[scale=0.4]{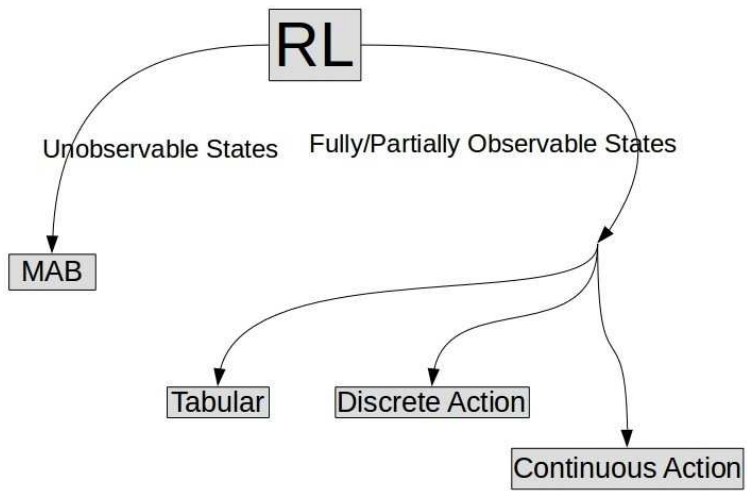}

\caption{General View of RL: Generally, RL is divided into two main groups.
In the fully/partially observable states group, the continuous action
space category is mainly concerned about the continuous control problem.}

\label{Fig: RL_figs}
\end{figure}

At the onset of RL development, most of the algorithms were concerned
with discrete state and action spaces, a.k.a Tabular Methods, see
Figure \ref{Fig: RL_figs}. For example, Dynamic Programming (DP)
methods \cite{bellman2013dynamic} were the frontiers in this development.
From a classical perspective, RL algorithms were developed in the
context of DP. Although these methods are well developed mathematically,
they need a complete and precise model of the environment \cite{sutton1998reinforcement}.
In order for an agent to be free from a complete knowledge about the
environment or even learn model of the environment, Monte Carlo (MC)
methods were introduced \cite{michie1968boxes}. These methods do
not need an explicit notion of the environment. Hence, they are more
flexible compared to DP methods. On the other hand, these methods
need a full episode experience of the agent-environment interaction.
This shortcoming makes MC methods incapable of learning in a step-by-step
incremental computation \cite{sutton1998reinforcement}. The Temporal-Difference
(TD) methods were introduced to both run the agent without a knowledge
of the environment and also stay away from full episode learning.
TD methods need no model and they can learn incrementally through
an agent-environment interaction \cite{sutton1998reinforcement}.
TD methods have furnished the next generation of RL algorithms. Some
important TD methods are Sarsa \cite{rummery1994line}, which is an
on-policy algorithm, and Q-Learning \cite{watkins1992q}, which is
an off-policy algorithm. Actor-critic methods were also introduced
initially as part of the TD methods \cite{witten1977adaptive}. However,
the actor-critic method became a backbone of the continuous control
RL methods. 

There is also another related problem in RL that addresses learning
in environments where the state is unknown. An $n$-armed bandit,
also known as Multi-Armed Bandits (MAB), problem \cite{sutton1998reinforcement}
is an analogy to a one-armed bandit slot machines. The difference
is that it has $n$ arms rather than one. The relationship between
$n$-armed bandits and RL is that we only have one state for the environment
(Figure \ref{Fig: RL_figs}). Not knowing the state can paradoxically
simplify and harden the problem. We will not address this type of
RL in this manuscript. In what comes next, we will focus on Q-learning
algorithm as it is the core element of this manuscript as well.

\subsection{Discrete Q-Learning Algorithms}

As mentioned before, the original Q-learning algorithm \cite{watkins1992q}
was initially developed to avoid learning a model of the environment.
This algorithm tries to solve (\ref{eq: Optimal_Q}) by updating $Q\left(s,a\right)$
through iterations of the form
\begin{equation}
Q\left(s,a\right)\leftarrow Q\left(s,a\right)+\alpha\left(r+\gamma\underset{a^{'}}{\max}Q\left(s^{'},a^{'}\right)-Q\left(s,a\right)\right),\label{eq: Q-Learning iteration}
\end{equation}
where $0<\alpha\leqslant1$ is a learning rate. The Q-learning algorithm
\cite{watkins1992q} usually works well in discrete state and action
spaces. An immediate generalization to the continuous state spaces
is to use a function estimator to approximate $Q\left(s,a\right)$.
However, this generalization is usually unstable or even divergent
\cite{tsitsiklis1997analysis}, when a nonlinear function estimator
is used. The good news is that this issue can be fixed by experience
replay and delayed updates of the function estimators \cite{mnih2015human}.

With the advances of the deep learning methods, in specific Deep Neural
Networks (DNN) \cite{goodfellow2016deep}, recently Mnih et al. have
been able to use Q-learning with a deep learning architecture to show
that a Deep Q-network (DQN) agent could achieve a human level performance
across a set of $49$ classic Atari $2600$ games \cite{mnih2015human}.
Furthermore, in a recent work by Hessel et al., it is shown that DQN
can even perform better if it is accompanied by combining other improvements
to the original algorithm \cite{hessel2017rainbow}. 

It is noteworthy to mention that the DQN algorithm takes in continuous
states and evaluates the action-value function for each discrete action
, see Algorithm \ref{Alg: DQN}. Hence, in case of a continuous action
space, finding 
\begin{equation}
a_{t}=\underset{a}{\arg\max}Q\left(\phi\left(s_{t}\right),a;\theta\right)\label{eq: Argmax_DQN}
\end{equation}
becomes challenging \cite{lillicrap2015continuous}. In fact, many
authors have tried to solve this maximization problem by introducing
function estimators that are more manageable than the non-linear neural
networks \cite{gaskett1999q,baird1993reinforcement}. However, their
approaches would in turn reduce the capacity of the function estimator
itself. 

\begin{algorithm}[H]
\begin{raggedright}
\textbf{Initialization.}
\par\end{raggedright}
\begin{raggedright}
Initialize replay memory $D$ to capacity $N$.
\par\end{raggedright}
\begin{raggedright}
Initialize action-value function $Q$ with random weights $\theta$.
\par\end{raggedright}
\begin{raggedright}
Initialize target action-value function $\hat{Q}$ with weights $\theta^{-}=\theta$.
\par\end{raggedright}
\begin{raggedright}
\textbf{Algorithm.} 
\par\end{raggedright}
\begin{raggedright}
For $episode=1\ldots M$ do
\par\end{raggedright}
\begin{itemize}
\item \begin{raggedright}
Initialize sequence $s_{1}$ and preprocessed sequence $\phi_{1}=\phi\left(s_{1}\right)$
\par\end{raggedright}
\item \begin{raggedright}
For $t=1\ldots T$ do 
\par\end{raggedright}
\begin{itemize}
\item \begin{raggedright}
$a_{t}=\left\{ \begin{array}{cc}
\underset{a}{\arg\max}Q\left(\phi\left(s_{t}\right),a;\theta\right) & \Pr=1-\varepsilon,\\
\sim U\left(\mathcal{A}\right) & \Pr=\varepsilon.
\end{array}\right.$
\par\end{raggedright}
\item \begin{raggedright}
Execute action $a_{t}$ and observe $r_{t}$ and $s_{t+1}$
\par\end{raggedright}
\item \begin{raggedright}
$\phi_{t+1}=\phi\left(s_{t+1}\right)$
\par\end{raggedright}
\item \begin{raggedright}
Store $\left(\phi_{t},a_{t},r_{t},\phi_{t+1}\right)$ in $D$
\par\end{raggedright}
\item \begin{raggedright}
Sample random minibatch of transitions $\left(\phi_{j},a_{j},r_{j},\phi_{j+1}\right)$
from $D$
\par\end{raggedright}
\item \begin{raggedright}
$y_{j}=\left\{ \begin{array}{cc}
r_{j} & episode\,terminates\,at\,j+1,\\
r_{j}+\gamma\underset{a^{'}}{\max}\hat{Q}\left(\phi_{j+1},a^{'};\theta^{-}\right) & otherwise.
\end{array}\right.$
\par\end{raggedright}
\item \begin{raggedright}
$\underset{\theta}{\min}\underset{j}{\sum}\left(y_{j}-Q\left(\phi_{j},a_{j};\theta\right)\right)^{2}$
\par\end{raggedright}
\item \begin{raggedright}
Every $C$ steps: $\theta^{-}=\theta$
\par\end{raggedright}
\end{itemize}
\end{itemize}
\caption{DQN with Experience Replay \cite{mnih2015human}: As a side note,
we need to mention that since the DQN with experience replay was designed
to read in raw pixels, a preprocessing function $\phi\left(.\right)$
has been used to alleviate the computational burden of the algorithm.
However, this preprocessing function can be substituted with a unity
function, specially if a deep architecture is not needed.}

\label{Alg: DQN}
\end{algorithm}

\subsection{Continuous Q-Learning Algorithms}

As we observed in the previous part, if the action space is continuous,
finding a solution to (\ref{eq: Argmax_DQN}) becomes computationally
expensive. Hence, to stay away from another computationally heavy
optimization in the original Q-learning algorithm, one quick way is
to make the whole action space discrete. This remedy would work for
low dimensional cases. However, the problem's complexity will grow
exponentially for higher dimensions. For example, if $m=\dim\left(\mathcal{A}\right)$,
and each coordinate is discretized in $k$ values, then there will
be a totall of $k^{m}$ actions. Even a 2-element coarse discretization
of $\left\{ -1,1\right\} $, for an action coordinate $\left[-1,1\right]$,
would result in $2^{m}$ actions. As $m$ grows, the problem becomes
intractable. As a result, a full and explicit discretization will
not work. This caveat is also mentioned by Duan et al. in \cite{duan2016benchmarking}.
Examples of this approach and its variants, that only work in low
dimensional cases, can be found in \cite{kobayashi2014q,bonarini2009reinforcement,millan2002continuous,takahashi1999continuous,fukao1998q}.

To circumvent this issue, the Normalized Advantage Function (NAF)
has been introduced as a Q-learning algorithm that acts in continuous
action spaces \cite{gu2017deep,gu2016continuous}. NAF uses a quadratic
advantage function to find appropriate actions at each state. The
parameters of this quadratic function are learned through the same
network that estimates the value function $V\left(s;\theta^{V}\right)$
and the advantage function $A\left(s,a;\theta^{V}\right)$. In fact,
the actions are not found through a complex optimization on the action-value
function $Q\left(s,a;\theta^{Q}\right)$, but a trivial optimization
of the advantage function $A\left(s,a;\theta^{V}\right)$. The main
issues with this novel method are that one has to estimate many different
functions to find the action and at the same time use a predefined
quadratic form for the advantage function $A\left(s,a;\theta^{V}\right)$.
We believe that this algorithm has a spirit of an actor-critic algorithm
rather than a Q-learning one. In Q-learning methods, only the action-value
function is learned.

\subsection{Other Continuous RL Algorithms}

The continuous action space RL resembles the continuous control problem
in control theory. Hence, the RL continuous action space problem is
also called continuous control. A natural method to solve the continuous
control problem in RL has been through using estimates of the policy.
In general, the Policy Gradient (PG) algorithms directly try to find
a policy $\pi_{\theta}\left(s,a\right)$, parameterized by $\theta$,
that maximizes the discounted reward $R_{t}$ or a surrogate of it.
In algorithms of type Q-learning, the action-value function $Q\left(s,a\right)$
is found first and in the second phase the policy $\pi\left(s,a\right)$
is derived from $Q\left(s,a\right)$. On the other hand, in PG algorithms,
the policy $\pi\left(s,a\right)$ is estimated directly. In the vanilla
version of the PG algorithm, a cost function is defined as 
\begin{equation}
J\left(\theta\right)=\mathbb{E}_{\tau\sim\Pr_{\theta}\left(\tau\right)}\left\{ \sum_{t=1}^{T}r_{t}\right\} ,\label{eq: Vanilla_PG_Cost}
\end{equation}
for a probability 
\begin{equation}
\Pr_{\theta}\left(\tau\right)=\Pr\left(s_{1}\right)\prod_{i=1}^{T-1}\mathcal{P}_{s_{i}s_{i+1}}^{a_{i}}\pi_{\theta}\left(s_{i},a_{i}\right)\label{eq: Trajectory_Prob}
\end{equation}
of the trajectory history $\tau=\left\{ s_{1},a_{1},\ldots,s_{T},a_{T}\right\} $.
Then the policy gradient of $\pi_{\theta}\left(s,a\right)$ is estimated
by
\begin{equation}
\nabla_{\theta}J\left(\theta\right)=\mathbb{E}_{\tau\sim\Pr_{\theta}\left(\tau\right)}\left\{ \left(\sum_{t=1}^{T-1}\nabla_{\theta}\ln\pi_{\theta}\left(s_{i},a_{i}\right)\right)\left(\sum_{t=1}^{T}r_{t}\right)\right\} ,\label{eq: PG}
\end{equation}
which later updates the parameter $\theta$ of the policy. From a
chronological perspective, REINFORCE \cite{williams1992simple} has
been the first PG method that maximizes the expected return by explicitly
manipulating the policy function $\pi_{\theta}\left(s,a\right)$.
Improvements to this algorithm was introduced through a class of algorithms
called Truncated Natural Policy Gradients (TNPG) \cite{kakade2002natural,peters2003policy,bagnell2003covariant,peters2008reinforcement}.

Recently, in the context of deep RL, Lillicrap et al. have proposed
a stable continuous control algorithm named as Deep Deterministic
Policy Gradient (DDPG) \cite{lillicrap2015continuous}. The algorithm
is based on the Deterministic Policy Gradient (DPG) algorithm \cite{silver2014deterministic}
that involves an actor function $\mu\left(s;\theta^{\mu}\right)$
and a critic, which is the action value function $Q\left(s,a;\theta^{Q}\right)$.
This algorithm is essentially performing a change of variables in
differention of $Q\left(s,a;\theta^{Q}\right)$ using $\mu\left(s;\theta^{\mu}\right)$. 

With the advance of the deep learning methods \cite{goodfellow2016deep},
methods like Trust Region Policy Optimization (TRPO) \cite{schulman2015trust},
and its generalization \cite{schulman2015high}, have shown faithful
results in addressing continuous control problem in high-dimensional
cases. Another recent algorithm, that can handle both continuous and
discrete action spaces, is the Proximal Policy Optimization (PPO)
\cite{schulman2017proximal}. Due to its relative simplicity comparred
to TRPO, it is getting an attention similar to DQN algorithms \cite{heess2017emergence}. 

Other authors have brought about methods that use a parameter model
of the policy function, but in a gradient-free context. Methods like
Cross Entropy Method (CEM) \cite{rubinstein1999cross,szita2006learning}
and Covariance Matrix Adaptation Evolution Strategy (CMA-ES) \cite{hansen2001completely}
are gradient-free algorithms that directly estimate the policy function
parameters. Policy optimization has also been addressed in the context
of Expectation-Maximization, namely Reward-Weighted Regression (RWR)
\cite{peters2007reinforcement,kober2009policy}. RWR methods are guaranteed
to converge to a local optimum solution. In line with RWR method,
Relative Entropy Policy Search (REPS) \cite{peters2010relative} uses
advanced optimization techniques to solve for policy function parameters.
A thorough performance analysis of these methods can be found in \cite{duan2016benchmarking}.

\subsection{Manuscript Organization and Main Contribution}

As we have seen so far, the continuous control problem has been a
challenge for Q-learning algorithms. In this paper, our main contribution
will concern about a simple trick to make the Q-learning algorithms
capable of handling continuous actions. We will show that how a linear-complexity
finite step modification can escape the curse of dimensionality and
make the Q-learning algorithm applicable on a discretized action space. 

In the next section, we will sketch the generalization of the the
Q-learning algorithm into the continuous action domain by introducing
the Finite Step Q-Learning (FSQ) algorithm. In the third section,
we will show some simulated examples to show the capability of our
method. Finally, we will wrap up the manuscript with a conclusion
and future work section. 

\section{Finite Step Q-Learning (FSQ).}

The original Q-learning algorithms, including DQN Algorithm \ref{Alg: DQN},
with a function estimator, e.g. a neural network, is taking a state
$s$ as an input and produces a vector of outputs for each discrete
action $a^{j}$, for $j=1\ldots m$ with $card\left(\mathcal{A}\right)=m$.
In other words, the function estimator $\boldsymbol{Q}\left(\boldsymbol{s},\boldsymbol{a};\theta\right)$
looks like
\begin{equation}
\boldsymbol{Q}\left(\boldsymbol{s},\boldsymbol{a};\theta\right)=\left(Q^{1}\left(\boldsymbol{s};\theta\right),\ldots,Q^{m}\left(\boldsymbol{s};\theta\right)\right)^{tr},\label{eq: Vector_Q}
\end{equation}
where $tr$ stands for a transform operator. So, there is a total
of $m$ discrete scalar action-value functions to be estimated. As
long as $m$ is relatively small, finding both the estimator $\boldsymbol{Q}\left(\boldsymbol{s},\boldsymbol{a};\theta\right)$
and finding
\begin{equation}
\boldsymbol{a}_{t}=\underset{\boldsymbol{a}}{\arg\max}Q\left(\boldsymbol{s}_{t},\boldsymbol{a};\theta\right)\label{eq: Crux}
\end{equation}
are computationally feasible. However, for a continuous space action,
using a Q-learning algorithm would encounter two different challenges.
\begin{enumerate}
\item If the action space is kept continuous, then one must solve (\ref{eq: Crux})
by a numerical method, like gradient ascent, through iteratively solving
\begin{equation}
\boldsymbol{a}^{\left(z+1\right)}=\boldsymbol{a}^{\left(z\right)}+\beta\nabla_{\boldsymbol{a}}Q\left(\boldsymbol{s}_{t},\boldsymbol{a}^{\left(z\right)};\theta\right),
\end{equation}
for some learning rate $\beta$. This makes the Q-learning algorithm
slow. So, from a computational and speed of convergence perspective,
this option is not advised.
\item If the action space is discretized, then the Q-learning algorithms
would not be able to work in high dimensional problems because of
the curse of dimensionality. As we stated before, if $m=\dim\left(\mathcal{A}\right)$,
and each coordinate is discretized in $k$ values, then there will
be an exponential totall of $k^{m}$ discretized actions. As a result,
a full and explicit discretization will not work if $m$ is relatively
large.
\end{enumerate}
In fact, we are going to show that one trick can salvage a discretized
version of Q-learning algorithms including DQN Algorithm \ref{Alg: DQN}.
In the next part, we will express our proposed Finite Step Q-learning
(FSQ) methodology that handles this issue.

\subsection{FSQ Methodology}

From a practical perspective, in reality, the actuators of a mechatronics
system \cite{ohnishi1996motion}, e.g. a robot, cannot immediately
move from an action $\boldsymbol{a}_{t}\in\mathcal{A}\left(s_{t}\right)$,
at time $t$, to any another action $\boldsymbol{a}_{t+1}\in\mathcal{A}\left(\boldsymbol{s}_{t+1}\right)$,
at the next time step $t+1$. We accept that a jump from an action
to any other action could happen in emulators and simulators. However,
it cannot happen in real systems. This fact can give us the idea that,
in a linear setting, two consequent actions are related to each other
by
\begin{equation}
\left|\boldsymbol{a}_{t+1}-\boldsymbol{a}_{t}\right|=\triangle\boldsymbol{a}.\label{eq: Disc_diff}
\end{equation}
Just to have a consistent notation, in (\ref{eq: Disc_diff}), the
absolute value is operated on all $m$ elements of $\boldsymbol{a}\in\mathcal{A}\subset\mathbb{R}^{m}$,
as follows 
\begin{equation}
\left|\boldsymbol{a}_{t+1}-\boldsymbol{a}_{t}\right|=\left(\left|a_{t+1}^{1}-a_{t}^{1}\right|,\ldots,\left|a_{t+1}^{m}-a_{t}^{m}\right|\right)^{tr}=\left(\triangle a^{1},\ldots,\triangle a^{m}\right)^{tr}=\triangle\boldsymbol{a}.
\end{equation}
In other words, for each element $j$ of the vector of continuous
actions $a\in\mathcal{A}\subset\mathbb{R}^{m}$, we can say
\begin{equation}
a_{t+1}^{j}=a_{t}^{j}+d_{t}^{j}\triangle a^{j},\,j=1\ldots m,\label{eq: Discr_diff_d}
\end{equation}
with $d^{j}\in\left\{ -1,0,1\right\} $. The vector $\boldsymbol{d}$
is simply defining the next action direction, in each element of $\boldsymbol{a}$;
namely right, left or nothing. In essence, using (\ref{eq: Discr_diff_d}),
we can simply consider the actions to be similar to states. Consequently,
it is only needed to learn the direction of each coordinate for the
$\boldsymbol{a}$ vector. In other words, the action-value function
estimator would take the form of 
\begin{equation}
\boldsymbol{Q}\left(\boldsymbol{v},\boldsymbol{d};\theta\right)=\left(Q^{1,1}\left(\boldsymbol{v};\theta\right),Q^{1,2}\left(\boldsymbol{v};\theta\right),Q^{1,3}\left(\boldsymbol{v};\theta\right),\ldots,Q^{m,1}\left(\boldsymbol{v};\theta\right),Q^{m,2}\left(\boldsymbol{v};\theta\right),Q^{m,3}\left(\boldsymbol{v};\theta\right)\right)^{tr}.\label{eq: Auxilary_Q_Estimator}
\end{equation}
Here, $\theta$ is the parameter of estimation for the function estimator
$\boldsymbol{Q}$. We call (\ref{eq: Auxilary_Q_Estimator}) the Auxiliary
Q-Function (AQF), $\boldsymbol{v}=\left(\boldsymbol{s},\boldsymbol{a}\right)$
the auxiliary state, and $\boldsymbol{d}$ the auxiliary action. 

Using this trick, the total number of elements in $\boldsymbol{Q}\left(\left(\boldsymbol{s},\boldsymbol{a}\right),\boldsymbol{d};\theta\right)$
will be $3m$. If a 3-element coarse discretization of the action
space was supposed to be implemented, there would be $3^{m}$ actions
instead. Hence, the AQF (\ref{eq: Auxilary_Q_Estimator}) has a linear
complexity in its output arguments. Since vector $\boldsymbol{d}$
is a vector of length $3m$, we can partitian both $\boldsymbol{d}$
and $\boldsymbol{Q}\left(\left(\boldsymbol{s},\boldsymbol{a}\right),\boldsymbol{d};\theta\right)$,
see (\ref{eq: Auxilary_Q_Estimator}), as follow
\begin{equation}
\boldsymbol{d}=\left(\begin{array}{c}
d^{1}\\
\vdots\\
d^{m}
\end{array}\right),\label{eq: Partitianed_d}
\end{equation}
\begin{equation}
\boldsymbol{Q}\left(\boldsymbol{v},\boldsymbol{d};\theta\right)=\left(\begin{array}{c}
\boldsymbol{Q}^{1}\left(\boldsymbol{v};\theta\right)\\
\vdots\\
\boldsymbol{Q}^{m}\left(\boldsymbol{v};\theta\right)
\end{array}\right),\label{eq: Partitianed_Auxilary_Q_Estimator}
\end{equation}
in which $\boldsymbol{Q}^{j}\left(\boldsymbol{v};\theta\right)\in\mathbb{R}^{3\times1}$
and $j=1\ldots m$. As a result, both the functions estimator (\ref{eq: Auxilary_Q_Estimator})
and solving the optimization 
\begin{equation}
d_{t}^{j}=\underset{k\in\left\{ 1,2,3\right\} }{\arg\max}\left(Q^{j,k}\left(\boldsymbol{v}_{t};\theta\right)-1\right),\,j=1\ldots m\label{eq: Auxilary_Q_max_Calc}
\end{equation}
become tractable from a computational perspective. 

\subsection{FSQ Algorithm}

We have presented the FSQ algorithm with experience replay in Algorithm
\ref{Alg: FSQ}. Another difference between the FSQ Algorithm \ref{Alg: FSQ}
and the DQN Algorithm \ref{Alg: DQN} comes from the target setup.
We need to keep in mind that since each $d^{j}$ is specifying a certain
direction for each element of $\boldsymbol{a}$, the target step of
the FSQ Algorithm \ref{Alg: FSQ} should become
\begin{equation}
y_{k}^{j,p_{k}^{j}}=\left\{ \begin{array}{cc}
r_{k} & episode\,terminates\,at\,k+1,\\
r_{k}+\gamma\max\hat{\boldsymbol{Q}}^{j}\left(\phi_{k};\theta^{-}\right) & otherwise,
\end{array}\right.\label{eq: Target_Setup}
\end{equation}
for $j=1\ldots m$ having $p_{k}^{j}=\arg\max\hat{\boldsymbol{Q}}^{j}\left(\phi_{k};\theta^{-}\right)$.
The vector of targets is then defined as 
\begin{equation}
\boldsymbol{y}=\left(\begin{array}{c}
\boldsymbol{y}^{1}\\
\vdots\\
\boldsymbol{y}^{m}
\end{array}\right)\label{eq: Targets}
\end{equation}
for $\boldsymbol{y}^{j}=\left(y^{j,1},y^{j,2},y^{j,3}\right)^{tr}\in\mathbb{R}^{3\times1},\,j=1\ldots m$.
Furthermore in Algorithm \ref{Alg: FSQ}, $\left\Vert .\right\Vert _{2}$
is the norm-2. If a neural network is used as the action-value function
estimator, $\boldsymbol{Q}\left(\left(\boldsymbol{s},\boldsymbol{a}\right),\boldsymbol{d};\theta\right)$
can be seen as a neural net ingesting vectors $\left(\boldsymbol{s},\boldsymbol{a}\right)$
and exporting $m$ different $3$-element vectors, as in (\ref{eq: Partitianed_Auxilary_Q_Estimator}). 

Finally, as a side note, the target update (\ref{eq: Target_Setup})
happens simultaneously for all $m$ partitioned vectors in (\ref{eq: Targets}).
This is also another difference to ordinary discrete action space
methods in which only one target element is updated. 

\begin{algorithm}[H]
\begin{raggedright}
\textbf{Initialization.}
\par\end{raggedright}
\begin{raggedright}
Initialize replay memory $D$ to capacity $N$.
\par\end{raggedright}
\begin{raggedright}
Initialize $\triangle\boldsymbol{a}$
\par\end{raggedright}
\begin{raggedright}
Initialize action-value function $\boldsymbol{Q}$ with random weights
$\theta$.
\par\end{raggedright}
\begin{raggedright}
Initialize target action-value function $\hat{\boldsymbol{Q}}$ with
weights $\theta^{-}=\theta$.
\par\end{raggedright}
\begin{raggedright}
\textbf{Algorithm.} 
\par\end{raggedright}
\begin{raggedright}
For $episode=1\ldots M$ do
\par\end{raggedright}
\begin{itemize}
\item \begin{raggedright}
Initialize sequence $\boldsymbol{s}_{1}$, $\boldsymbol{a}_{1}=\boldsymbol{0}$
and preprocessed sequence $\phi_{1}=\phi\left(\boldsymbol{s}_{1},\boldsymbol{a}_{1}\right)$
\par\end{raggedright}
\item \begin{raggedright}
For $t=1\ldots T$ do 
\par\end{raggedright}
\begin{itemize}
\item \begin{raggedright}
For $j=1\ldots m$ do 
\par\end{raggedright}
\begin{itemize}
\item \begin{raggedright}
$d_{t}^{j}=\left\{ \begin{array}{cc}
\underset{\boldsymbol{d}^{j}}{\arg\max}\left(\boldsymbol{Q}\left(\phi\left(\boldsymbol{s}_{t},\boldsymbol{a}_{t}\right),\boldsymbol{d}^{j};\theta\right)-1\right) & \Pr=1-\varepsilon,\\
\sim U\left(\left\{ -1,0,1\right\} \right) & \Pr=\varepsilon.
\end{array}\right.$
\par\end{raggedright}
\item \begin{raggedright}
$a_{t+1}^{j}=a_{t}^{j}+d_{t}^{j}\triangle a^{j}$
\par\end{raggedright}
\end{itemize}
\item \begin{raggedright}
Execute action $\boldsymbol{a}_{t}$ and observe $r_{t}$ and $\boldsymbol{s}_{t+1}$
\par\end{raggedright}
\item \begin{raggedright}
$\phi_{t+1}=\phi\left(\boldsymbol{s}_{t+1}\right)$
\par\end{raggedright}
\item \begin{raggedright}
Store $\left(\phi_{t},\boldsymbol{d}_{t},r_{t},\phi_{t+1}\right)$
in $D$
\par\end{raggedright}
\item \begin{raggedright}
Sample random minibatch $\mathcal{B}$ of transitions $\left(\phi_{k},\boldsymbol{d}_{k},r_{k},\phi_{k+1}\right)$
from $D$
\par\end{raggedright}
\begin{itemize}
\item \begin{raggedright}
For $j=1\ldots m$ do
\par\end{raggedright}
\begin{itemize}
\item \begin{raggedright}
$y_{k}^{j,p_{k}^{j}}=\left\{ \begin{array}{cc}
r_{k} & episode\,terminates\,at\,k+1,\\
r_{k}+\gamma\max\hat{\boldsymbol{Q}}^{j}\left(\phi_{k};\theta^{-}\right) & otherwise,
\end{array}\right.$
\par\end{raggedright}
\end{itemize}
\item \begin{raggedright}
$\underset{\theta}{\min}\left\Vert \boldsymbol{y}-\boldsymbol{Q}\left(\phi,\boldsymbol{d};\theta\right)\right\Vert _{2}$
for the minibatch $\mathcal{B}$
\par\end{raggedright}
\end{itemize}
\item \begin{raggedright}
Every $C$ steps: $\theta^{-}=\theta$
\par\end{raggedright}
\end{itemize}
\end{itemize}
\caption{FSQ with Experience Replay: $\phi\left(.\right)$ is a preprocessing
function that can be substituted with a unity function if a deep architecture
is not needed.}

\label{Alg: FSQ}
\end{algorithm}

\subsection{Some Notes on the Convergence of the FSQ Method}

Although our method breaks the curse of dimensionality in its type
of discretization, we need to mention some hidden increases in computational
complexity introduced as part of the FSQ formulation. In particular,
in FSQ, the \textquotedblleft state\textquotedblright{} $\left(s,a\right)$
in the Q-function encompasses both the original state $s$ and the
action $a$ at each time step. The new action space is now much smaller,
but the state space has effectively increased by the dimensionality
of the original action space. So that will increase the time/sample
complexity of finding a function approximation for Q. How much it
increases seems to depend on the assumptions made about $\boldsymbol{Q}\left(\left(\boldsymbol{s},\boldsymbol{a}\right),\boldsymbol{d};\theta\right)$.
On the other hand, the convergence properties of the gradient descent
methods depend on the Lipschitz conditions of the functions being
optimized \cite{karimi2016linear}. From a practical perspective,
the Lipschitz constants do not increase exponentially with respect
to the increase in dimensionality of a neural network family function.
So, FSQ is possibly an affordable trade-off by an increase in the
number of iterations to find a good approximation for $\boldsymbol{Q}\left(\left(\boldsymbol{s},\boldsymbol{a}\right),\boldsymbol{d};\theta\right)$
in reducing the size of the action space.

\section{Example}

\begin{figure}[H]
\includegraphics[scale=0.4]{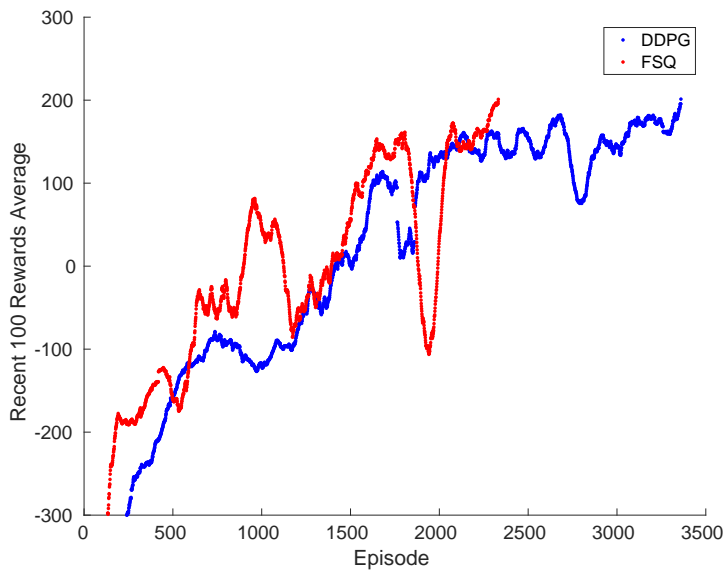}

\caption{FSQ compared to DDPG}

\label{Fig: FSQvsDDPG}
\end{figure}

In Figure \ref{Fig: FSQvsDDPG}, we are comparing the FSQ method with
DDPG \cite{lillicrap2015continuous} algorithm. The FSQ implementation
used in this example uses Algorithm \ref{Alg: FSQ} equipped with
a variant of the Prioritized Experience Replay (PER) introduced in
\cite{schaul2015prioritized}, and double Q-learning explained in
\cite{hasselt2010double}.

We used the OpenAI Gym \cite{1606.01540} 'LunarLanderContinuous-v2'
environment. This environment has a continuous state space of dimension
$8$ and a continuous action space of dimension $2$. The DDPG implementation
was picked from the OpenAI Baseline algorithms repository \cite{baselines}.
We put the criteria for algorithm success when an average reward of
at least $200$ over the last $100$ consecutive episodes is reached.
We used the default hyperparameters provided by the DDPG Baseline
code \cite{baselines}. In our implementation, we used the hyperparameters
mentioned in Table \ref{Table: FSQ_hyperparameters}. The function
estimator for the FSQ algorithm was a shallow neural network with
$128$ neurons. We used Adam method for the neural network optimization
algorithm \cite{kingma2014adam}.

As we can see in Figure \ref{Fig: FSQvsDDPG}, our repeated experiments
show that the FSQ Algorithm \ref{Alg: FSQ} reaches the milestone
in less than $\sim1000$ episodes ahead of the DDPG method. It is
also of interest to mention that we used a rough increment $\triangle a^{j}=0.75,\,j=1,2$,
in our implementation. Even with such a value, when the action space
domain is $\left[-1,1\right]^{2}$, the FSQ algorithm can achieve
an acceptable learning curve as shown in Figure \ref{Fig: FSQvsDDPG}.

Finally, if we wanted to use a $3$-element coarse discretization
of $\left\{ -1,0,1\right\} $, for the action space $\left[-1,1\right]$$^{2}$,
the discretized action space could have $3^{2}$ elements. However,
in our implementation, for this environment, we only have $3\times2$.

\section{Conclusion and Future Work}

Q-learning \cite{watkins1992q} has been a cornerstone of RL in the
past few decades. However, its extension, even in deep learning realm,
has been confined to discrete action spaces \cite{mnih2015human}.
In this article, we have been able to systematically introduce an
algorithm, see FSQ Algorithm \ref{Alg: FSQ}, to bring in Q-learning
into the continuous action space domain. This extension can ease up
the way to use Q-learning in continuous control cases by introducing
a linear-complexity discretization in the action space. At the same
time, the FSQ method, Algorithm \ref{Alg: FSQ}, is unifying the state
and action spaces. It is noteworthy to mention that the type of discretization
introduced in this article is not confined to Q-learning methods alone.
In fact, this linear-complexity discretization can be used in any
RL algorithm in which it is originally designed for a discrete action
space. 

In Algorithm \ref{Alg: FSQ}, we have furnished the the general form
of the FSQ. However, in the example, presented in this article, we
have only considered a shallow neural network for a relatively low
dimensional state space problem. In a follow up work, we will try
to show the capabilities of our algorithm in a context that the raw
signal would be the only input to our algorithm. Hence, a DNN architecture
would be used to estimate the action-value function $Q$. Furthermore,
the discretization value $\triangle\boldsymbol{a}$ can have effects
on the stability of our method. We believe that $\triangle\boldsymbol{a}$
must have a lower bound to provide stability and convergence. In our
follow up work, we will try to address this point as well.

\section{Acknowledgment}

This research was carried out at the Jet Propulsion Laboratory, California
Institute of Technology, under a contract with the National Aeronautics
and Space Administration.

\copyright{} 2018 California Institute of Technology. Government
sponsorship acknowledged.

\section{Figures}

\section{Tables}

\begin{table}[H]
\begin{tabular}{|c|c|}
\hline 
Hyperparameter & Value\tabularnewline
\hline 
Memory Size  & $50000$\tabularnewline
\hline 
Target Network Update Interval & $1000$\tabularnewline
\hline 
Memory Batch Size & $32$\tabularnewline
\hline 
Neural Batch Size & $32$\tabularnewline
\hline 
Optimizer & $Adam$\cite{kingma2014adam}\tabularnewline
\hline 
Learning Rate & $0.0005$\tabularnewline
\hline 
Discount Rate & $0.99$\tabularnewline
\hline 
$\varepsilon_{max}$ & $1.0$\tabularnewline
\hline 
$\varepsilon_{min}$ & $0.1$\tabularnewline
\hline 
$\varepsilon$-Decay Rate & $0.001$\tabularnewline
\hline 
$\triangle a^{j},\,j=1,2$ & $0.75$\tabularnewline
\hline 
\end{tabular} 

\caption{FSQ Example Hyperparameters}

\label{Table: FSQ_hyperparameters}
\end{table}

\bibliographystyle{plain}

\end{document}